\documentclass[11pt]{article}
\usepackage{coling2020}
\usepackage{times}
\usepackage{url}
\usepackage{latexsym}

\usepackage{multirow}
\usepackage{graphicx}
\usepackage{subfigure}
\usepackage{tabularx}
\usepackage{hhline}
\usepackage{array}
\usepackage{amssymb}
\usepackage{amsfonts}
\usepackage{amsmath}
\usepackage{mathrsfs}
\usepackage{color}
\usepackage{booktabs}
\usepackage{amssymb}
\usepackage[T1]{fontenc}

\makeatletter
\def\thanks#1{\protected@xdef\@thanks{\@thanks
        \protect\footnotetext{#1}}}
\makeatother

\colingfinalcopy % Uncomment this line for the final submission

\title{R4: A Framework for Route Representation and Route Recommendation}

\author{Ran Cheng$^*$\thanks{*These authors contribute equally.} \\
  {\tt \small{chengran.cr@alibaba-inc.com}} \\\And
  Chao Chen$^*$ \\
  {\tt \small{cc201598@alibaba-inc.com}} \\\And
  Longfei Xu$^*$ \\
  {\tt \small{longfei.xl@alibaba-inc.com}} \\\AND
  Shen Li \\
  {\tt \small{jingfu.ls@alibaba-inc.com}} \\\And
  Lei Wang \\
  {\tt \small{shiluo.wl@alibaba-inc.com}} \\\And
  Hengbin Cui \\
  {\tt \small{alexcui.chb@alibaba-inc.com}} \\\AND
  Kaikui Liu \\
  {\tt \small{damon@alibaba-inc.com}} \\\And
  Xiaolong Li \\
  {\tt \small{xl.li@alibaba-inc.com}} \\\AND
  Alibaba Group
}

\date{}

\begin{document}
\maketitle

\hspace{0.5em}

\begin{abstract}
  Route recommendation is significant in navigation service. Two major challenges for route recommendation are route representation and user representation. Different from items that can be identified by unique IDs in traditional recommendation, routes are combinations of links (i.e., a road segment and its following action like turning left) and the number of combinations could be close to infinite. Besides, the representation of a route changes under different scenarios. These facts result in severe sparsity of routes, which increases the difficulty of route representation. Moreover, link attribute deficiencies and errors affect preciseness of route representation. Because of the sparsity of routes, the interaction data between users and routes are also sparse. This makes it not easy to acquire user representation from historical user-item interactions as traditional recommendations do. To address these issues, we propose a novel learning framework R4. In R4, we design a sparse $\&$ dense network to obtain representations of routes. The sparse unit learns link ID embeddings and aggregates them to represent a route, which captures implicit route characteristics and subsequently alleviates problems caused by link attribute deficiencies and errors. The dense unit extracts implicit local features of routes from link attributes. For user representation, we utilize a series of historical navigation to extract user preference. R4 achieves remarkable performance in both offline and online experiments. 
\end{abstract}

% \keywords{route representation, route recommendation, route navigation}

\section{Introduction} \label{section:introduction}

\begin{figure}[htbp] 
  \centering    %居中
   
  \subfigure %第一张子图
  {
    \begin{minipage}[t]{0.47\linewidth}
    \centering          %子图居中
    \includegraphics[width=\linewidth]{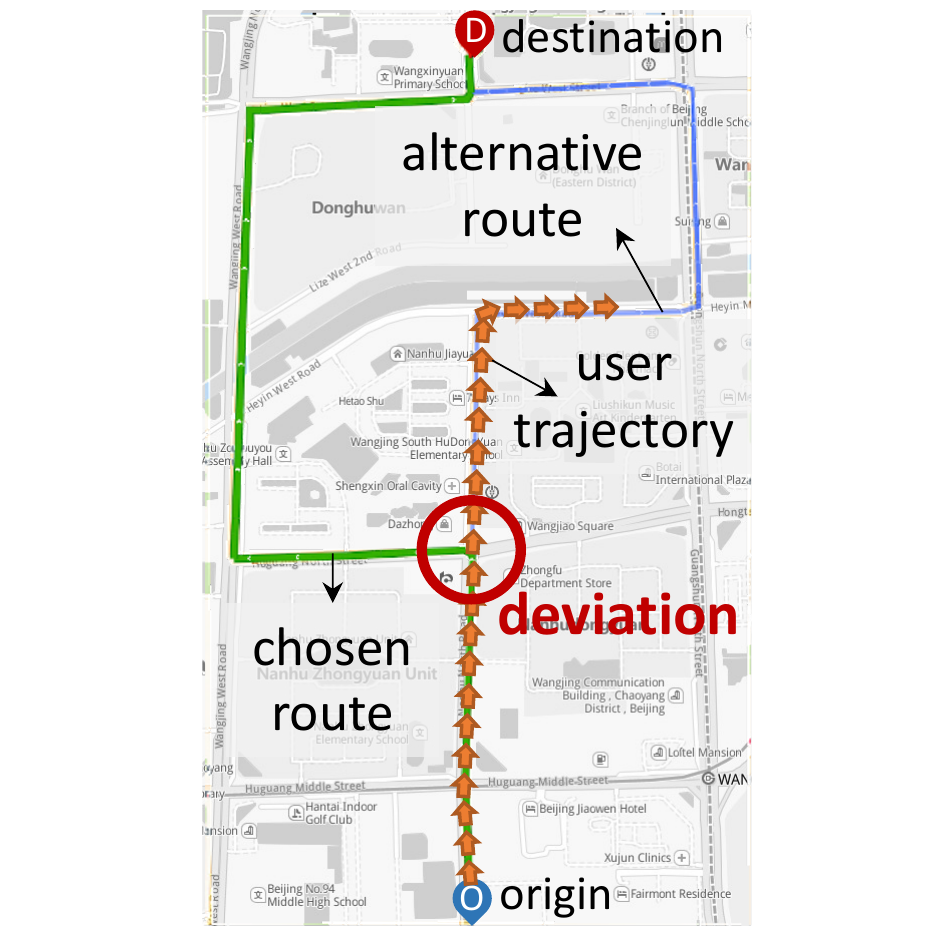}   %以pic.jpg的0.5倍大小输出
    \end{minipage}
  }
  \subfigure %第二张子图
  {
    \begin{minipage}[t]{0.47\linewidth}
    \centering      %子图居中
    \includegraphics[width=\linewidth]{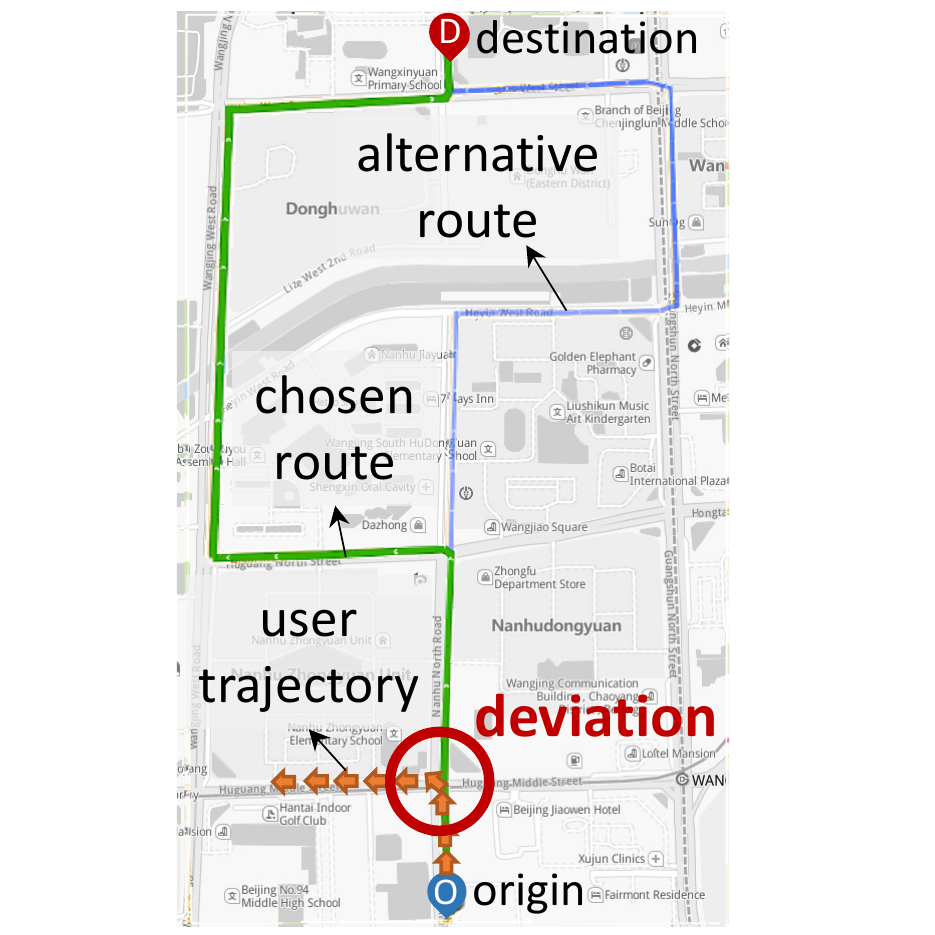}   %以pic.jpg的0.5倍大小输出
    \end{minipage}
  }
  \caption{\textbf{Deviation.} As long as the user trajectory deviates from the chosen route, it will be regarded as a deviation, whether the user deviates to an alternative route (left) or a route which is not displayed (right).} %  %大图名称
  \label{figure:deviation}  %图片引用标记
\end{figure}

Route recommendation plays an important role in navigation service, since it affects travel experience of users. When a user requests a route planning from an origin to a destination, navigation service will gather hundreds of available routes and then select several of them to recommend to the user. The user chooses one route and enjoys his travel accompanying with the navigation service. If the user deviates from the chosen route as shown in Figure \ref{figure:deviation}, it implies user dissatisfaction with the route to a certain extent. Since user satisfaction is determined by route characteristics and user preference, the representations of route and user are two vital parts of route recommendation.

The sparsity of routes makes route representation a challenging task. It is not practical to assign a unique ID to each route as other recommendation systems do to an item, such as YouTube recommendations \cite{covington2016deep}, Wide $\&$ Deep framework \cite{cheng2016wide} and DIN \cite{zhou2018deep}. The reason is that routes are combinations of links and the number of combinations could be close to infinite, where \textbf{link} is defined as the concatenation of a road segment and its following action (e.g., going forward and turning left). Besides, the representations of a route under different scenarios, such as different traffic conditions, could be different. These facts make routes severely sparse, enlarging difficulty of learning route representation.

Moreover, sufficient and accurate link attributes are significant for route representation. However, some route characteristics cannot be depicted by existing statistical features. For example, if there are a lot of cars parked on both sides of a road, it is not very easy to drive through it. Finding a feature to describe such characteristic is a complicated task. Besides, data errors occasionally appear in link attributes. For instance, a blocked road is mistakenly identified as a passable road. These problems also make it challenging to acquire precise route representation.

In addition to route representation, user representation is also pivotal in route recommendation. Traditional recommendations utilize historical user-item interactions to extract user preference for items. For example, if a user frequently clicks on a set of items recently, there is a high probability that he will buy one of them. However, as aforementioned, routes are severely sparse. It implies that the interactions between users and routes are sparse as well. Hence, it is not feasible to learn user representation from historical user-route interactions.

To address above issues, we propose a novel framework R4. In addition to route-level features, R4 introduces link-level features to obtain route representation. A sparse $\&$ dense network is designed to generate two types of route representations. The sparse unit learns link ID embeddings and aggregates them to represent a route. It is capable of discovering implicit characteristics of the route, which subsequently alleviates problems caused by link attribute deficiencies and errors. For the dense part, we refer to the ideas of residual network \cite{he2016deep} and sentence CNN \cite{rakhlin2016convolutional}. A deep residual network is applied to capturing different levels of route local features (i.e., detours) and then aggregating them to represent a route. With regard to user representation, a series of user historical navigation are utilized to extract user preference. After that, DCN-V2 \cite{wang2020dcn} is adopted to generate combination features from two types of route representations and user representation. R4 achieves superior performance on both offline and online experiments.

We carry out an analysis on link embeddings and find out that they contain information of link static attributes, which validates the effectiveness of route representation in R4. Furthermore, a similarity unit is designed to expose hidden characteristics of links. It learns similarity scores between link ID embeddings and vectors of link static attributes. A smaller similarity value implies more hidden information contained in the ID embedding of a link. By conducting analysis on links with small similarity values, we discover some hidden features, which proves that R4 has the ability to capture implicit characteristics in addition to static attributes of links. Meanwhile, these hidden features could be helpful in discovering new attributes or correcting data errors in link attributes. This procedure can provide more sufficient and accurate link static features, which in return benefits route representation learning.

Thus, the contribution of this paper is four-fold:

\begin{itemize}
  \item A learning framework R4 is proposed for route representation and route recommendation, and it achieves remarkable performance on both offline and online experiments.
  \item To the best of our knowledge, we are the first to introduce link IDs in route representation.
  \item The sparse $\&$ dense network proposed in this paper represents a route from both link IDs and link attributes, which can be applied to route representation in any scenario.
  \item The similarity unit is designed to expose implicit characteristics of links, which is helpful in finding new attributes and correcting attribute errors, and in return, provides sufficient and accurate link data for route representation learning.
\end{itemize}

\section{Related Work} \label{section:relatedwork}

In route recommendation, most of classical works focus on improving route searching by optimizing link costs. Various features of links, such as distance, travel time, gas consumption and personalized information, have been introduced to optimize link costs \cite{tian2009finding,andersen2013ecotour,kanoulas2006finding,dai2015personalized,funke2015personalized}. However, in these works, the cost of the route generally is a simple linear accumulation of its link costs, which leads to poor route representation because the interaction information between links that compose the route cannot be captured. Some people implement more complicated route cost functions like using neural networks to learn cost functions of the A* algorithm \cite{wang2019empowering}. However, this method could not directly leverage the global information of a route since it is only based on the links that have been predicted and guesses the rest of the route by reinforcement learning. Besides, costs of links cannot be calculated in parallel in this work because the costs of candidate links at each step rely on the result of the previous step, which makes the method unable to be applied to a large road network. In order to learn better route representation while considering the feasibility for application in large road network, we propose a two-step approach for route recommendation: 1. searching routes as many as possible between two points in a road network; 2. ranking these routes. Existing multi-route algorithms \cite{abraham2013alternative,bader2011alternative,delling2017customizable} have already been capable of searching hundreds of routes between an origin and a destination. Therefore, our work focuses on route ranking. In this strategy, route representation could be learned more effectively and efficiently than the previous works.

If a route is regarded as an item, then route ranking task is actually an item recommendation task. Unlike traditional items, routes are sequences composed of different number of links, which makes routes much sparser than items. Therefore, it is impractical to represent routes by IDs which are widely used in item recommendation. As a result, ID-based algorithms, such as xDeepFM \cite{lian2018xdeepfm}, AutoInt \cite{song2019autoint}, Sampling-Bias-Corrected neural modeling \cite{yi2019sampling} and YouTube recommendation model \cite{covington2016deep}, cannot handle route representation. Besides, it also leads to the infeasibility of constructing the ID-based user-item interactions which have been proved to be effective in obtaining user representation, e.g. DIN \cite{zhou2018deep} and DIEN \cite{zhou2019deep}.

Compared with item representation in traditional recommendation, route representation is closer to sentence representation in natural language processing (NLP), and a link is to a route what a word is to a sentence. Word embedding and sentence embedding are basic and mature researches in NLP. Many works have achieved outstanding results, such as Skip-Gram \cite{mikolov2013efficient}, Transformer \cite{vaswani2017attention}, BERT \cite{devlin-etal-2019-bert}, etc. Nevertheless, these models cannot be directly transferred to the route representation task because of the following two aspects: 1. there are hundreds of millions of links in a large road network, which are much more and sparser than words; 2. the states of links change more frequently than grammatical and semantic meanings of words which are stable in different time. Therefore, to tackle these problems, we propose a framework R4 to gain more suitable representations of routes and users.

\begin{table*}[htbp]
\centering 
  \setlength{\belowcaptionskip}{0.3cm}
  \begin{small}
  \begin{tabular}{c|c|l|l|l}
  \hline
  Domain                    & Notation    & \multicolumn{1}{c|}{Feature} & \multicolumn{1}{c|}{\begin{tabular}[c]{@{}c@{}}Dimensionality\\  or Range\end{tabular}} & \multicolumn{1}{c}{Type} \\ \hline
  \multirow{3}{*}{context}  & \multirow{3}{*}{$\boldsymbol{f_c}$}      & time when the route recommendation was initiated                & $\sim$10          & discrete              \\
                            &    & spherical distance between an origin and a destination          & (0, 1000{]} km                               & continuous      \\
                            &    & …                            & …                                            & …                    \\ \hline
  \multirow{3}{*}{user}     & \multirow{3}{*}{$\boldsymbol{f_u}$} & age        & $\sim$10\textasciicircum{}2        & discrete                \\
                            &    & number of historical deviations     & (0, 500{]}                                   & continuous       \\
                            &    & …                            & …                                            & …            \\ \hline
  \multirow{3}{*}{route}    & \multirow{3}{*}{$\boldsymbol{f_r}$}    & route distance     & (0, 2000{]}  km         & continuous               \\
                            &    & route ETA       & (0, 1000{]} min                            & continuous            \\
                            &    & …                            & …                                            & …                          \\ \hline
  \multirow{1}{*}{link id}    & \multirow{1}{*}{$p_{id}$}  & link ID        & $\sim$10\textasciicircum{}8      & discrete        \\ \hline
  \multirow{3}{*}{link static} & \multirow{3}{*}{$\boldsymbol{l_s}$}  & number of lanes     & $\sim$10    & discrete        \\
                               & & link length                   & (0,2]  km                                    & continuous        \\
                               & & …                             & …        & …        \\ \hline
  \multirow{3}{*}{link dynamic} & \multirow{3}{*}{$\boldsymbol{l_d}$} & link traffic condition                   & $\sim$10      & discrete     \\
                               & & link ETA                     & (0,120] min     & continuous        \\
                               & & …                             & …                                            & …          \\ \hline
  \multirow{2}{*}{link position} & \multirow{2}{*}{$\boldsymbol{l_p}$} & cumulative distance from the origin to the link     & (0, 2000{]} km     & continuous      \\
                               & & …                            & …                                            & …             \\ \hline
  \end{tabular}
  \end{small}
  \caption{Feature Description} 
  \label{table:feature}
\end{table*}

\section{Methodology}

\subsection{Background}

Item recommendation systems in E-commerce normally collect thousands of candidate items first, and then rank them with a more sophisticated model in order to obtain a better performance of recommendation. Similar with that, we adopt a  two-step approach to tackle with route recommendation task. First, we generates up to hundreds of candidate routes using CRP algorithm \cite{delling2017customizable} which is able to be applied to realtime navigation system. After that, we rank candidate routes and recommend the top one. Route ranking is what we concentrate on in this paper.

Since traffic condition changes frequently, especially in densely populated cities like Beijing, simply ranking shortest path ahead does not work. Besides, user preference differs from person to person. For example, some prefer major roads despite of congested traffic, while others prefer minor roads with free traffic. These facts make it difficult to find a route-related indicator to measure  quality of a route. Therefore, we propose to measure route quality by user deviation behaviors. On the one hand, user deviation on the chosen route generally means that the user is not satisfied with it. In some cases, deviations indeed cannot represent user preferences, such as user changing the destination. Such cases account for a very low proportion, and can be easily identified and filtered out. On the other hand, in addition to the route chosen by the user, the navigation service also provides multiple alternative routes. Therefore, if the user does not deviate from the chosen route, it means that the user prefers to it or thinks there is no problem with it, otherwise the user will deviate to other routes. 

In traditional recommendations, the conversion rate (CVR) is the probability that a user buys an item after clicking on it. Similarly, deviation rate (DR) is proposed to measure the probability that a user deviates from a route after choosing it. The difference is that traditional recommendation aims to maximize CVR, while route recommendation aims to minimize DR. Therefore, the target of R4 is to predict DR of candidate routes, and then recommend the route with the lowest DR to users.

\subsection{Feature Description}

Features for predicting DR in R4 are listed in Table \ref{table:feature}, categorized into 7 domains. 

Since route condition and user preference vary in different environment, context features $\boldsymbol{f_c}$ (e.g., the time when the route recommendation was initiated, spherical distance between an origin and a destination, route feature difference between the sample and other candidate routes) are involved. 

User features $\boldsymbol{f_u}$ refer to user profile, such as age, number of historical deviations. These features are helpful in DR prediction because personality traits influence user behaviors and different users may end up with different outcomes in similar circumstances. Besides, the number of historical deviations alleviates the bias of user on following recommended route.

Route features are denoted as $\boldsymbol{f_r}$ , including route distance, route ETA (i.e., estimated time of arrival), distance of road segments with congested traffic, etc. They are high correlated to user behavior during navigation. Other than route-level features, link-level features are also included to provide more detailed information. Link static features $\boldsymbol{l_s}$ represent relatively stable information of link like link length, while link dynamic features $\boldsymbol{l_d}$ refer to high-frequency changing information like link conditions. Link position features $\boldsymbol{l_p}$ reflect the offset of a link in a route, such as the cumulative distance from the origin to the link. For convenience, link ID is also regarded as a domain which is denoted as $p_{id}$.

It should be noted that some domains contain discrete features. In embedding layer of R4, these discrete features are projected to dense vectors using embedding technique and concatenated with the normalized continuous features. Embedding technique is a normal operation in NLP to transform high-dimensional sparse vector into low-dimensional dense vector. Discrete features will first be encoded as one-hot vectors. After that, embedding layer maps them into dense vectors. For example, with one-hot encoding, link ID becomes a one-hot vector with dimension of over 5 million. After that, the embedding layer maps it into a 32-dimension vector.

Besides, the predicted DR is denoted by $\hat y$ and the label $y$ is set to $1$ if the user deviates from the route, and $0$ otherwise.

\subsection{Overall Architecture}

\begin{figure*}[htbp]
  \centering
  \includegraphics[width=1\linewidth]{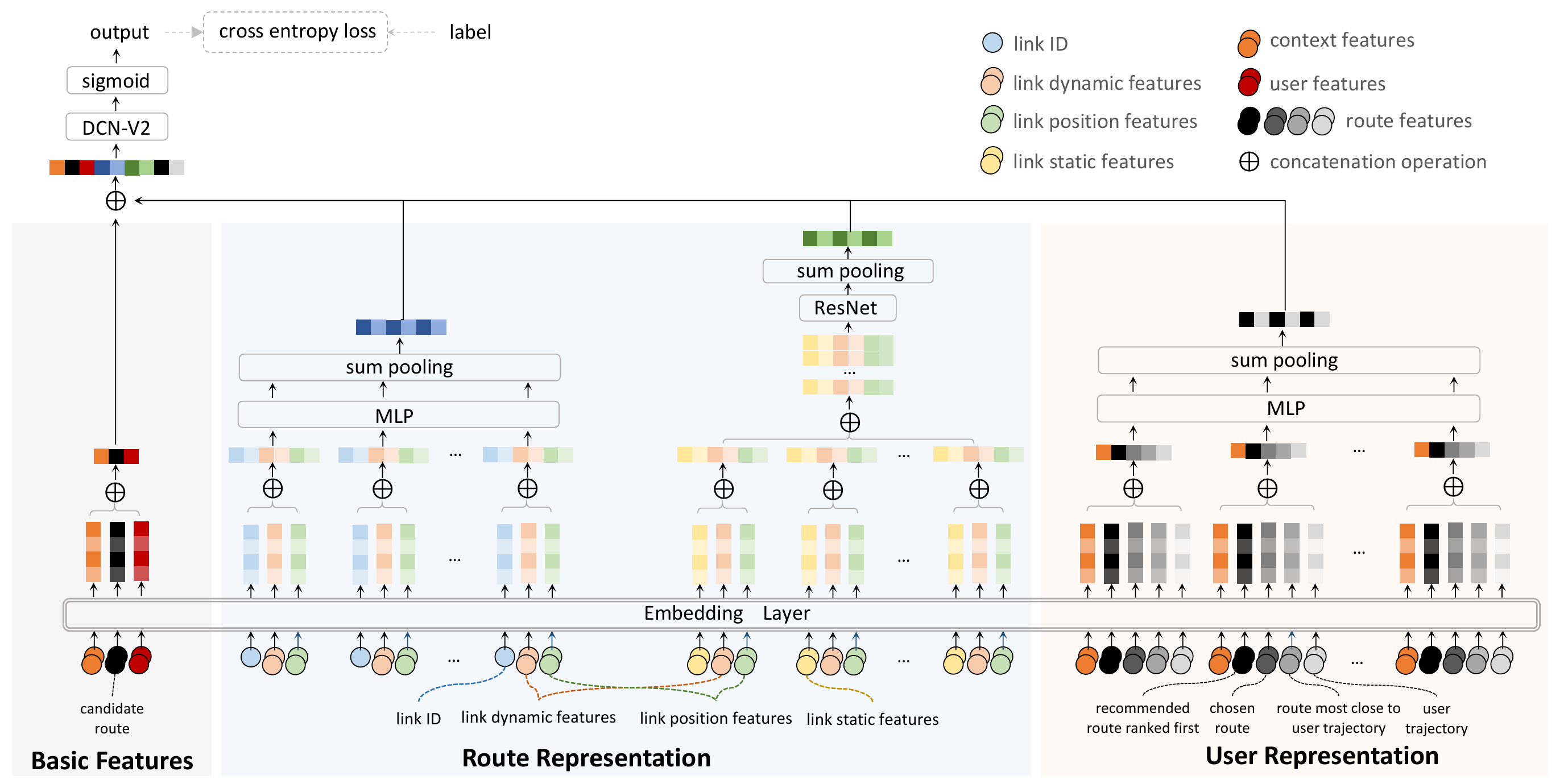}
  \caption{
    \textbf{Overall architecture of R4.} Route representation module utilizes link-level information to obtain route representation. User representation module learns user preference from information of a series of user historical navigation. The output of each module and basic features are concatenated and sent into DCN-V2 \cite{wang2020dcn}. Finally, the predicted DR is output after sigmoid function. In embedding layer, discrete features are projected to dense vectors and concatenated with the normalized continuous features.
  }
  \label{figure:overallarchitecture}
\end{figure*}

The overall architecture of R4 is presented in Figure \ref{figure:overallarchitecture}. R4 has two main modules, route representation module and user representation module. Their outputs are combined with the basic features to predict DR. Each module will be explained in the following sections.

\subsection{Base Model} \label{section:basemodel}
With basic features as input, the base model adopts DCN-V2 \cite{wang2020dcn} for DR predicting. As shown in Figure \ref{figure:overallarchitecture}, basic features are composed of context features $\boldsymbol{f_c}$, route features $\boldsymbol{f_r}$, and user features $\boldsymbol{f_u}$. DCN-V2 is a state-of-the-art model for feature crossing. The loss function of the base model is the cross-entropy loss defined as below:

\begin{equation}\label{equ:cross_entropy}
  L = - \frac {1} {|\boldsymbol D|} \sum_{\boldsymbol D} (y\log \hat y + (1-y)\log (1-\hat y)),
\end{equation}

where $|\boldsymbol D|$ is the size of the training set $\boldsymbol D$.

\subsection{Route Representation}

\subsubsection{Sparse Network}

This module aims to obtain route representation from link-level features. Three feature vectors of link are utilized: dense vector $\boldsymbol{l_e}$ of link ID $p_{id}$, vector $\boldsymbol{l_d}$ of link dynamic features, and vector $\boldsymbol{l_p}$ of link position features. 

Since a route is composed of multiple links, the sparse network learns link embeddings and aggregates them to represent a route. Referring to the solution in NLP, we replace low-frequency links with the unknown token.

Besides, a link could have different representations in different scenarios and positions. For example, the characteristics of the same link under free and congested traffic conditions are different. For another example, if a link includes a private road, the representation when it appears at the very beginning of a route is different from the representation when it appears in the middle of a route. Because the former is passable since that is the place where user starts navigation, the latter is probably not. Therefore, we introduce link dynamic features and link position features for link embedding learning.

The route representation module in Figure \ref{figure:overallarchitecture} demonstrates the approach to represent a route. If a route is composed of $M$ links, then we get a sequence of inputs $[\boldsymbol{f_s^1}, \boldsymbol{f_s^2},...,\boldsymbol{f_s^M}]$, where $\boldsymbol{f_s^i} = \boldsymbol{l_e^i} \oplus \boldsymbol{l_d^i} \oplus \boldsymbol{l_p^i}$ denotes the feature vector of the $i$-th link and $\oplus$ represents concatenation operation. To obtain route representation, a multi-layer perceptron (MLP) is applied to each $\boldsymbol{f_s}$ to acquire link embeddings. After that, a sum pooling layer adds up all link embeddings into route embedding.

Due to the introduction of link ID, the sparse network is capable of extracting implicit characteristics of routes, which cannot be easily depicted by existing statistical features.

\subsubsection{Dense Network}

\begin{figure*}[htbp]
  \setlength{\abovecaptionskip}{0.2cm}
  \setlength{\belowcaptionskip}{0.cm}
  \centering    %居中
  \subfigure[convolute along link sequence] %第一张子图
  {
    \begin{minipage}[t]{0.4\linewidth}
    \centering          %子图居中
    \includegraphics[width=\linewidth]{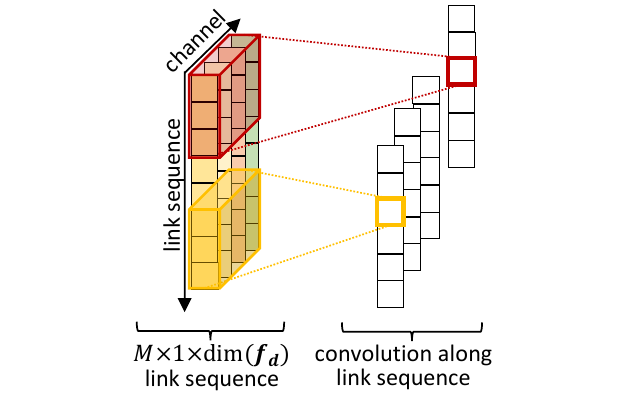}   %以pic.jpg的0.5倍大小输出
    \end{minipage}
    \label{figure:linkcnn}
  }
  \subfigure[ResNet20 captures information of 13 links] %第二张子图
  {
    \begin{minipage}[t]{0.55\linewidth}
    \centering      %子图居中
    \includegraphics[width=\linewidth]{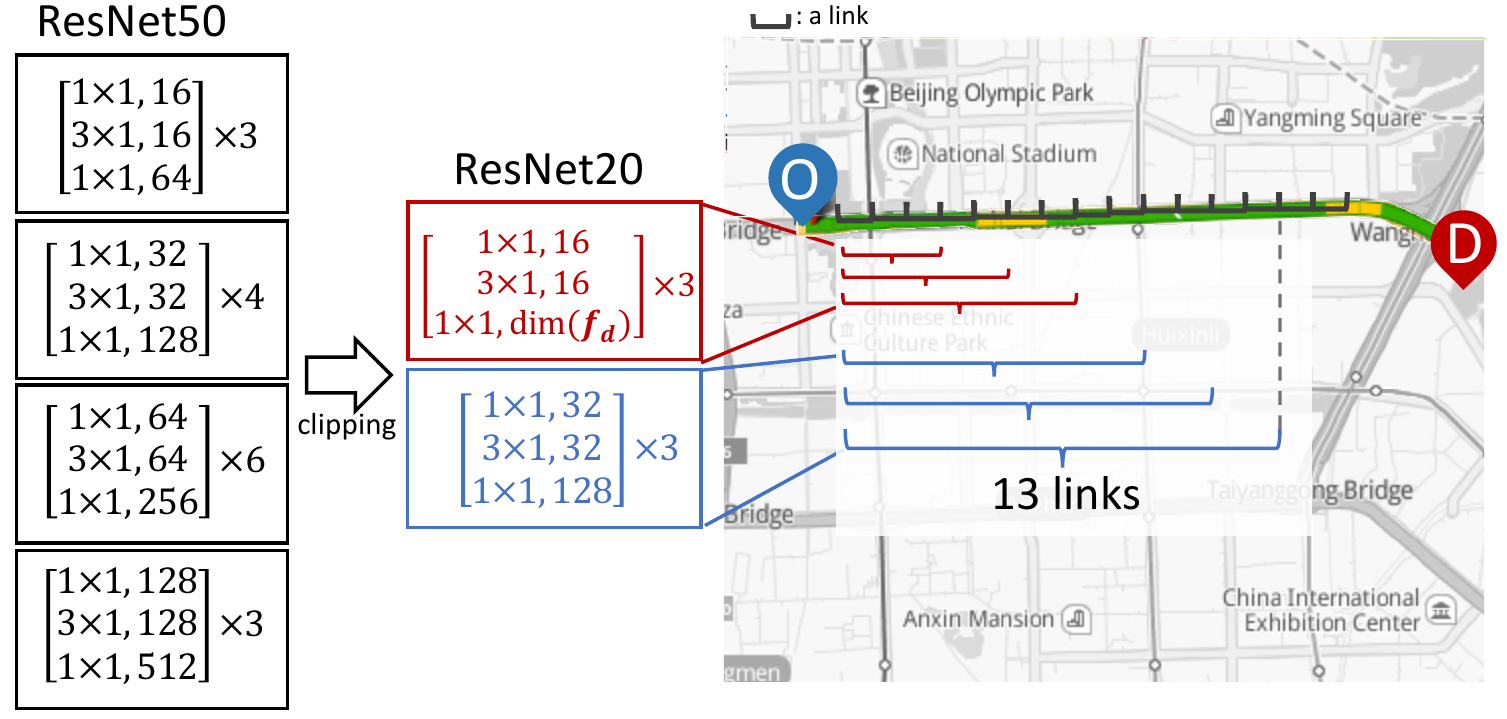}   %以pic.jpg的0.5倍大小输出
    \end{minipage}
    \label{figure:resnet}
  }
   
  \caption{Sequence Residual CNN Unit} %  %大图名称
  \label{figure:convolution}  %图片引用标记
\end{figure*}

The sparse network acquires global characteristics of routes. In contrast, the dense network aims at learning local characteristics of routes (e.g., detours). It employs link static features $\boldsymbol{l_s}$, link dynamic features $\boldsymbol{l_d}$ and link position attributes $\boldsymbol{l_p}$ to generate route representation.

The dense network adopts ResNet \cite{he2016deep} to extract local characteristics of routes. If a route is composed of $M$ links, then we get a sequence of inputs $\boldsymbol{S} = [\boldsymbol{f_d^1}, \boldsymbol{f_d^2},...,\boldsymbol{f_d^M}]$, where $\boldsymbol{f_d^i} = \boldsymbol{l_s^i} \oplus \boldsymbol{l_d^i} \oplus \boldsymbol{l_p^i}$ is the feature vector of the $i$-th link. Before input into ResNet, $\boldsymbol{S}$ is reshaped into a matrix with size $M  \times 1 \times \mathrm{dim}(\boldsymbol{f_d}) $, where $\mathrm{dim}(\boldsymbol{f_d})$ is the feature dimension of $\boldsymbol{f_d}$. Figure \ref{figure:linkcnn} illustrates the convolutional operation on the matrix, where the number of links is regarded as the height and the number of features as the channel. After multiple ResNet blocks, route representation is acquired by summing up each feature map.

\begin{figure}[htbp]
  \setlength{\abovecaptionskip}{0.2cm}
  \setlength{\belowcaptionskip}{0.cm}
  \centering
  \includegraphics[width=0.6\linewidth]{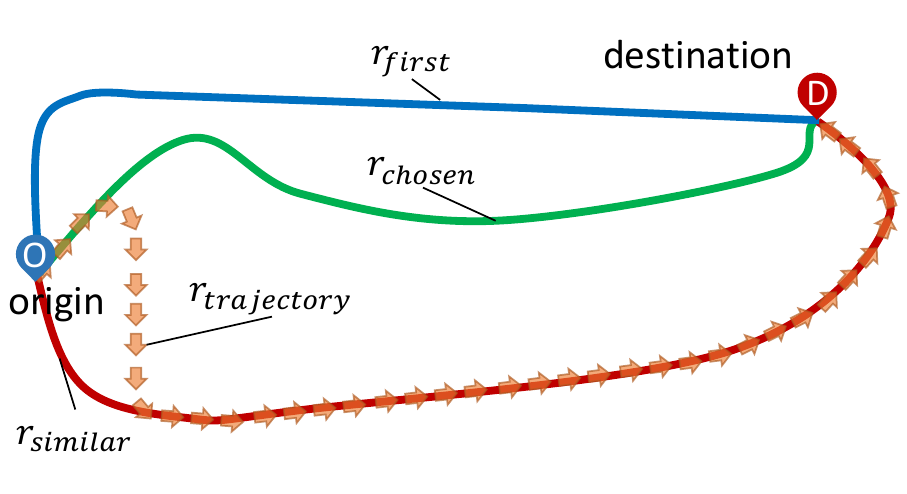}
  \caption{\textbf{Navigation record example.}}
  \label{figure:user_choice}
\end{figure}

\subsection{User Representation} \label{section:user representation}

This module extracts user preference from a sequence of user historical navigation. Figure \ref{figure:user_choice} gives an example of navigation record. Route $r_{first}$ is the first recommended route. The user chooses anther route to start navigation, which is labeled as $r_{chosen}$. However, the user deviates from the chosen route during navigation. Route $r_{trajectory}$ represents user trajectory. There is a candidate route which is the most similar with user trajectory, labeled as $r_{similar}$. If a user has $T$ navigation records recently, the input can be represented by a list $\boldsymbol{E} = [\boldsymbol{f_e^1}, \boldsymbol{f_e^2},..., \boldsymbol{f_e^T}]$, where $\boldsymbol{f_e^i} = \boldsymbol{f_c^i} \oplus \boldsymbol{f_{r_{first}}^i} \oplus \boldsymbol{f_{r_{chosen}}^i} \oplus \boldsymbol{f_{r_{similar}}^i} \oplus \boldsymbol{f_{r_{trajectory}}^i} $ is the feature vector of $i$-th navigation. $\boldsymbol{f_c^i}$ is the context information of navigation. $\boldsymbol{f_{r_{first}}^i}$ is the feature vector of the recommended route ranked first. $\boldsymbol{f_{r_{chosen}}^i}$ is the feature vector of user chosen route. $\boldsymbol{f_{r_{similar}}^i}$ is the feature vector of the planned route most close to user trajectory. $\boldsymbol{f_{r_{trajectory}}^i}$ is the feature vector of user trajectory. Features involved are listed in Table \ref{table:feature}. As shown in Figure \ref{figure:overallarchitecture}, an MLP generates feature combinations of $\boldsymbol{E}$ for each $\boldsymbol{f_e}$, and a sum pooling layer adds them up to obtain user representation.

\section{Experiment} \label{section:experiment}

\subsection{Datasets and Experiment Setup} 

\begin{table*}[htbp]
\centering
  \begin{scriptsize}
  \begin{tabular}{c|c|c|c|c|c|c|c|c}
  \hline
  & \begin{tabular}[c]{@{}c@{}}navigation\\ number\end{tabular} & \begin{tabular}[c]{@{}c@{}}user\\ number\end{tabular} & \begin{tabular}[c]{@{}c@{}}average number of\\ candidate routes \end{tabular} & \begin{tabular}[c]{@{}c@{}}average number of user\\ historical navigation\end{tabular} & \begin{tabular}[c]{@{}c@{}}number\\ of links\end{tabular} & \begin{tabular}[c]{@{}c@{}}number of\\ links visited\end{tabular} & \begin{tabular}[c]{@{}c@{}}average visited\\ frequency of link\end{tabular} & DR \\ \hline
  training set & 29,684 k & 5,234 k & 19 & 59 & 5,667 k & 2,479 k & 1,200 & 33\% \\ \hline
  \end{tabular}
  \end{scriptsize}
  \setlength{\belowcaptionskip}{0.3cm}
  \caption{Dataset Description}
  \label{table:dataset}
\end{table*}

The dataset used in experiments is provided by Amap, a leading navigation service provider in China. Each data sample is a navigation record of an anonymous user. Origins and destinations in the data are both at Beijing. It includes information of navigation context, candidate routes (i.e., routes that participate in recommendation), user profile, interactions between the user and recommended routes (e.g., choosing a route and deviation from a route), and user historical navigation records within last 90 days. Data samples that the user deviates from his chosen route for personal intention (e.g., changing destination and going for refuel) are filtered. Statistics of the dataset are summarized in Table \ref{table:dataset}. There are 29,684 k navigation records, belonging to 5,234 k users. The average number of candidate routes is 19, and the maximum number of candidate routes is 150. The average number of user historical navigation records is 59, the latest 30 records are utilized for model training. Among 5,667 k links in Beijing, 2,479 links are covered by user chosen routes in the dataset. The average frequency of links is 1,200. We use data of the last day for testing, and other 30 days for training.

Link IDs with frequency under 5 are replaced by the unknown token. The mini-batch size is set to be 512 and we use Adam \cite{kingma2014adam} as the optimizer. Besides, we adopt exponential decay technique, where the initial learning rate is 0.001 and the decay rate is 0.9.

\subsection{Methods} \label{section:methods}
Methods compared in experiments are listed as follows:

\begin{itemize}
  \item \textbf{Base Model.} We use DCN-V2 \cite{wang2020dcn} as the baseline model, as demonstrated in Section \ref{section:basemodel}. The DCN-V2 is a stacked combination of 2-layer cross network and 128-64 deep network (i.e., two hidden layers with outputs of size 128 and 64).
  \item \textbf{R4.} It is the framework proposed in this paper. In sparse network, the dimension of embedded link ID $\boldsymbol{l_e}$ is 32 and the dimension of route embedding is 64. Dense network uses ResNet50, as shown in Figure \ref{figure:resnet}. User representation module uses MLP with one hidden layer of size 64.
  \item \textbf{R4-WoS.} It differs from R4 by removing the sparse network.
  \item \textbf{R4-WoU.} It differs from R4 by removing the user representation module.
  \item \textbf{R4-WoD.} It differs from R4 by removing the dense network.
  \item \textbf{R4-WoB.} It differs from R4 by removing route features $\boldsymbol{f_r}$ from basic features.
  \item \textbf{R4-WoDP.} It differs from R4 by removing link dynamic features $\boldsymbol{l_d}$ and link position features $\boldsymbol{l_p}$ from the route representation module.
  \item \textbf{R4-WoD.} It differs from R4 by removing link dynamic features $\boldsymbol{l_d}$ from the route representation module.
  \item \textbf{R4-WoP.} It differs from R4 by removing link position features $\boldsymbol{l_p}$ from the route representation module.
  \item \textbf{R4-C (clipped).} To speed up model inference, the dense network uses ResNet20 instead of ResNet50 in online experiment, as displayed in Figure \ref{figure:resnet}.
\end{itemize}

\begin{table}[htbp] 
  \setlength{\belowcaptionskip}{0.3cm}
  \centering
  \begin{tabular}{c|c|c|c}
  \hline
                                                                                    & Model Name                & AUC  & Best Setting                 \\ \hline
  \multirow{5}{*}{\begin{tabular}[c]{@{}c@{}}different\\ modules\end{tabular}}        & Base Model                 & 0.7333    & $l$=2, {[}128, 64{]}           \\
                                                                                    & R4-WoU                   & 0.7595   & $e$=32, $h$=64, ResNet50         \\
                                                                                    & R4-WoD                   & 0.7564   & $e$=32, $h$=64           \\
                                                                                    & R4-WoS                   & 0.7442     & ResNet50            \\
 & R4-WoB              & 0.7600    & $e$=32, $h$=64, ResNet50 \\ \hline
  \multirow{3}{*}{\begin{tabular}[c]{@{}c@{}}different \\ link features\end{tabular}} & R4-WoDP           & 0.7578   & $e$=32, $h$=64, ResNet50 \\
                                                                                    & R4-WoP & 0.7600   & $e$=32, $h$=64, ResNet50\\
                                                                                    & R4-WoD         & 0.7576  & $e$=32, $h$=64, ResNet50 \\ \hline
  model clipping                                                                    & R4-C              & 0.7603  & $e$=32, $h$=64, ResNet20 \\ \hline
  best model                                                                        & R4                     & 0.7606  & $e$=32, $h$=64, ResNet50 \\ \hline
  \end{tabular}
  \caption{\textbf{Performance.} $l$ is the number of layers of cross network in DCN-V2 \cite{wang2020dcn} and [128, 64] is the hidden layer sizes of dense network in DCN-V2. $e$ and $h$ are the dimension of link ID embedding and route embedding respectively.}
  \label{table:comparison}
\end{table}

\subsection{Evaluation Metric}
We adopt AUC as the metric to measure the performance of DR prediction.
It reflects the probability that a model ranks a randomly chosen positive
sample higher than a randomly chosen negative sample, which is defined as
\begin{equation}
  AUC= \frac 1{|D^+|\cdot|D^-|} \sum_{D^+} \sum_{D^-} (\mathrm{I}({\hat y}^+ > {\hat y}^-)),
\end{equation}
where $D^+$ is the dataset of positive samples, $D^-$ is the dataset of negative samples, ${\hat y}^+$ is model output of a positive sample, ${\hat y}^-$ is model output of a negative sample, and $\mathrm{I}$ is indicator function.

Based on our historical experience, every 1$\%$ increases in offline AUC 
brings about 1$\%$ decreases in online DR.

\begin{table*}[htbp]
  \setlength{\belowcaptionskip}{0.3cm}
  \centering
  \begin{tabular}{c|c||c|c|c|c|c|c|c}
  \hline
    & DR & PV & ETA & route distance & toll & traffic light & turns &  highway  \\ \hline
    SD & 0.4648 & 79,157 & 2,115 s & 19.309 km & 2.03 ¥ & 14.09 & 12.15 & 4.207 km   \\ \hline
    ST & 0.4094 & 79,969 & 1,914 s & 21.007 km & 3.24 ¥ & 9.6 & 12.28 & 6.645 km   \\ \hline
    ST-H & 0.4042 & 81,454 & 1,975 s & 20.244 km & 2.91 ¥ & 10.3 & 11.80 & 6,032 km  \\ \hline
  Base Model & 0.3863 & 81,008 & 1,955 s & 20.537 km & 2.66 ¥ & 10.51 & 11.89 & 5.746 km   \\ \hline
  R4-C & 0.3622 & 79,115  & 1,966 s & 20.491 km & 2.67 ¥ & 10.36 & 11.48 & 5.600 km  \\ \hline
  \end{tabular}
  \caption{Performance Online}  %表格标题
  \label{table:online}
\end{table*}

\subsection{Experiments on Different Modules} 

\begin{figure}[htbp] \label{figure:clustering}  %图片引用标记
  \setlength{\abovecaptionskip}{0.2cm}
  \setlength{\belowcaptionskip}{0.cm}
  \centering    %居中
   
  \subfigure[link ID embedding clustering] %第一张子图
  {
    \begin{minipage}[t]{0.87\linewidth}
    \centering          %子图居中
    \includegraphics[width=\linewidth]{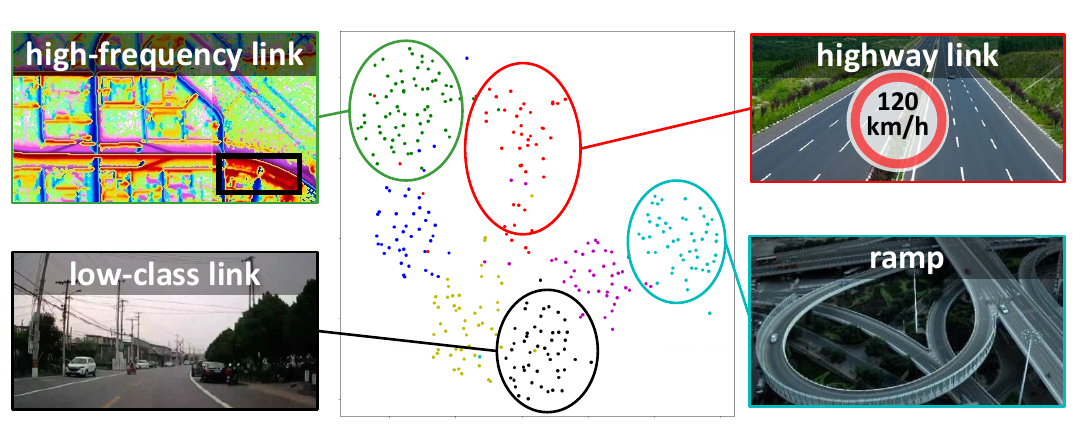}   %以pic.jpg的0.5倍大小输出
    \end{minipage}
    \label{figure:linkembedding}
  }
  \subfigure[user embedding clustering] %第二张子图
  {
    \begin{minipage}[t]{0.876\linewidth}
    \centering      %子图居中
    \includegraphics[width=\linewidth]{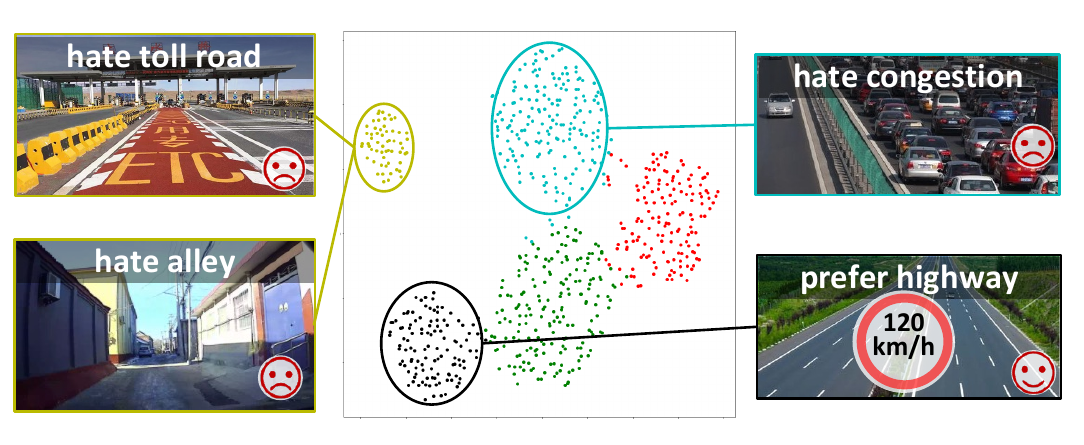}   %以pic.jpg的0.5倍大小输出
    \end{minipage}
    \label{figure:userembedding}
  }
  \caption{\textbf{Embedding Clustering.} We use k-means to cluster embedding vectors and present them after dimensionality reduction by t-SNE \cite{van2008visualizing}. Low-frequency links and users are removed. Each cluster has at least one characteristic very different from others. For example, most links in the red cluster of link ID embeddings are highways, so we assign it a highway label. (a) shows several clusters of link ID embeddings with labels such as slip road, highway, etc. and (b) shows several clusters of user embeddings with labels such as hating toll roads, preferring highways, etc.} %  %大图名称
\end{figure}

Results of experiments are shown in Figure \ref{table:comparison}. R4 improves AUC by 2.8$\%$ compared with the base model, which is a remarkable improvement. It indicates the effectiveness of R4 on route and user representation learnings.

The comparison between R4 and R4-WoS shows that the sparse network improves AUC by 1.6$\%$. In order to depict link ID embeddings learned by the sparse network, we adopt k-means to cluster them and t-SNE \cite{van2008visualizing} to project them into two-dimensional vectors. As displayed in Figure \ref{figure:linkembedding}, each cluster has one characteristic quite different from other clusters, such as slip road, highway, etc. It implies that link ID embeddings capture useful information.

Compared with R4-WoD, R4 improves AUC by 0.4$\%$, which manifests the effectiveness of the dense network. To prove that the dense network indeed has the ability to capture local properties of routes, we remove route features $\boldsymbol{f_r}$, including local features, from basic features to obtain R4-WoB. The similar performance between R4-WoB and R4 proves that the dense network is capable of covering all the information provided by $\boldsymbol{f_r}$. Furthermore, R4-WoB has better performance than R4-WoD. These two facts indicate that not only the dense network has learned route local features explicitly in basic features, but also stores other route characteristics implicitly.

R4 improves AUC by 0.08$\%$, compared with R4-WoU. Moreover, as shown in Figure \ref{figure:userembedding}, user embeddings after clustering reveal different user preferences, such as hating toll roads, hating congestion, etc. It demonstrates that user embeddings obtained from the historical navigation sequence can well represent user preferences.

\begin{figure}[htbp]
  \setlength{\abovecaptionskip}{0.2cm}
  \setlength{\belowcaptionskip}{0.cm}
  \centering
  \includegraphics[width=0.8\linewidth]{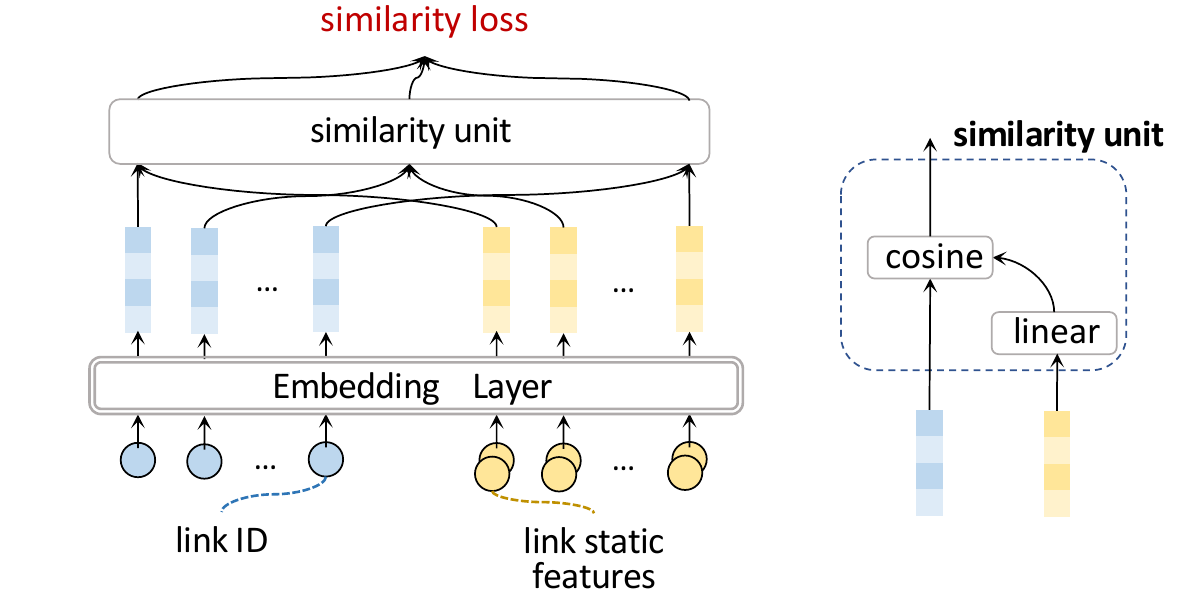}
  \caption{\textbf{Similarity unit.}}
  \label{figure:similarity_unit}
\end{figure}

Furthermore, we propose a similarity unit to reveal hidden features learned by the sparse network, as shown in Figure \ref{figure:similarity_unit}. First, we freeze all parameters in R4, and then calculate similarity between link ID embeddings and link static attributes and maximize it. Specifically, a linear transformation changes link static attributes $\boldsymbol{l_s}$ into a vector ${\boldsymbol{l_{s_t}}}$ which has the same size with the dense vector $\boldsymbol{l_e}$ of link ID. Next, we calculate cosine similarity between $\boldsymbol{l_e}$ and ${\boldsymbol{l_{s_t}}}$. One thing should be noticed is that cosine distance can only measure similarity between non-zero vectors. Hence, we add non-zero bias to avoid zero vectors as proposed in \cite{luo2018cosine}. In this regard, a smaller similarity value implies more extra information contained in the ID embedding of a link. We select links with small similarity values and conduct analysis on them to expose implicit link properties, including new attributes and attribute errors. The similarity loss is defined as:
\begin{equation}\label{equ:similarity}
  L_{sim} = - \frac {1}{|\boldsymbol {D_l}|} \sum_{\boldsymbol{D_l}}
  (\log \frac {1}{1+e^{-\boldsymbol{\mathrm{cosine}}({\boldsymbol{l_e}},{\boldsymbol{l_{s_t}}})}}) ,
\end{equation}
where $\boldsymbol{D_l}$ denotes links in the training set $\boldsymbol D$. Unknown tokens and zero-paddings do not participate in similarity loss computation.

\begin{figure}[htbp]
  \setlength{\abovecaptionskip}{0.2cm}
  \setlength{\belowcaptionskip}{0.cm}
  \centering    %居中
  \subfigure[width errors] %第一张子图
  {
    \begin{minipage}[t]{0.3\linewidth}
    \centering          %子图居中
    \includegraphics[width=\linewidth]{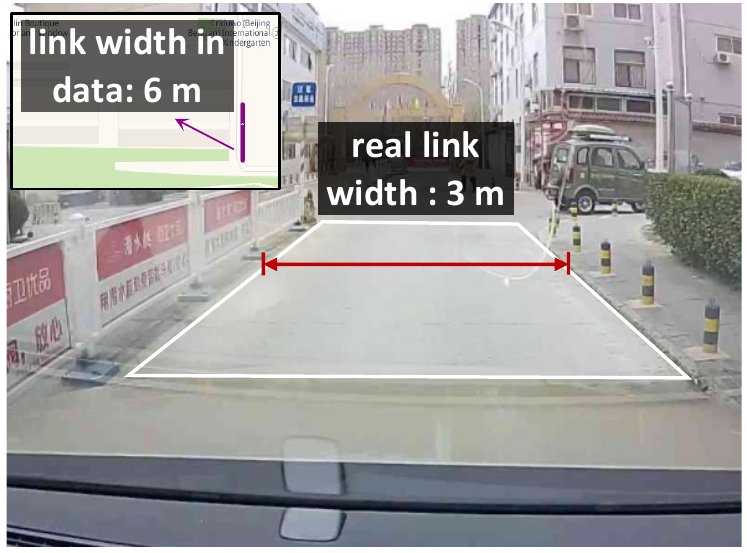}   %以pic.jpg的0.5倍大小输出
    \end{minipage}
    \label{figure:data_error_1}
  }
  \subfigure[blocked links] %第二张子图
  {
    \begin{minipage}[t]{0.3\linewidth}
    \centering      %子图居中
    \includegraphics[width=\linewidth]{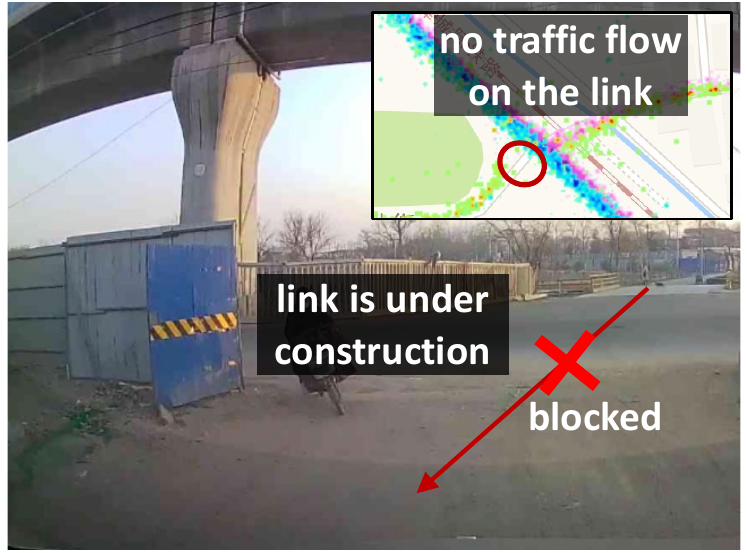}   %以pic.jpg的0.5倍大小输出
    \end{minipage}
    \label{figure:data_error_2}
  }
  \subfigure[hidden attributes] %第三张子图
  {
    \begin{minipage}[t]{0.3\linewidth}
    \centering      %子图居中
    \includegraphics[width=\linewidth]{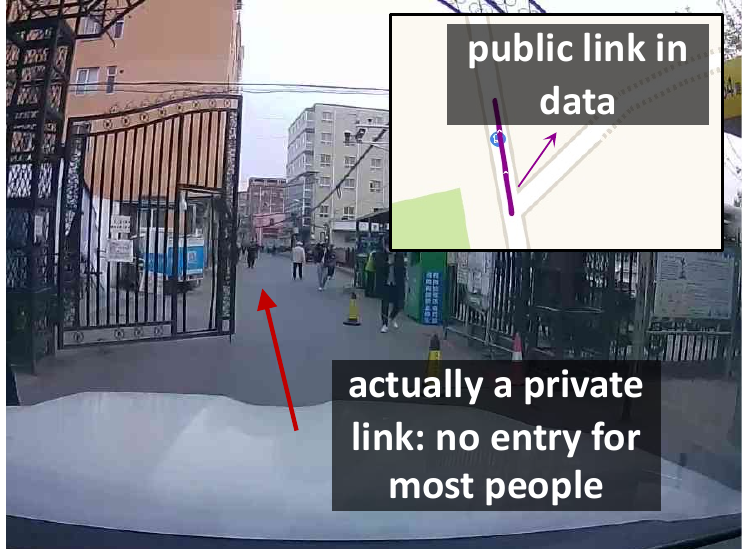}   %以pic.jpg的0.5倍大小输出
    \end{minipage}
    \label{figure:data_error_3}
  }
   
  \caption{\textbf{Cases of hidden feature detection.} } %  %大图名称
  \label{figure:data_error}  %图片引用标记
\end{figure}

Since the similarity unit tries to force a link ID embedding to imitate its corresponding vector of static attributes, if the similarity still keeps small, it probably means that the embedding has some hidden characteristics which have not been covered by static attributes, such as new features and data errors. We conduct analysis on links with small similarity values and find out cases given in Figure \ref{figure:data_error}. Figure \ref{figure:data_error_1} is a link whose width is 3 meters, but 6 meters in data; Figure \ref{figure:data_error_2} is a blocked link under construction, but stored as a passable link in data mistakenly. Figure \ref{figure:data_error_3} shows a private link of a village where most people cannot enter, but existing attributes cannot depict this situation explicitly. On the one hand, these cases are helpful to prove that link ID embeddings contain implicit characteristics of links. 

\subsection{Experiments on Different Link Features}

Table \ref{table:comparison} presents experiment results of ablating different link features. Compared with R4-WoD, R4 brings a big improvement of $0.3\%$ AUC. Compared with R4-WoP, R4 increases AUC by 0.06$\%$. These two observations confirm that the representations of a link should be different in different scenarios and positions.

\begin{figure*}[htbp]
  \setlength{\abovecaptionskip}{0.2cm}
  \setlength{\belowcaptionskip}{0.cm}
  \centering    %居中
  \subfigure[SD and ST-H] %第一张子图
  {
    \begin{minipage}[t]{0.3\linewidth}
    \centering          %子图居中
    \includegraphics[width=\linewidth]{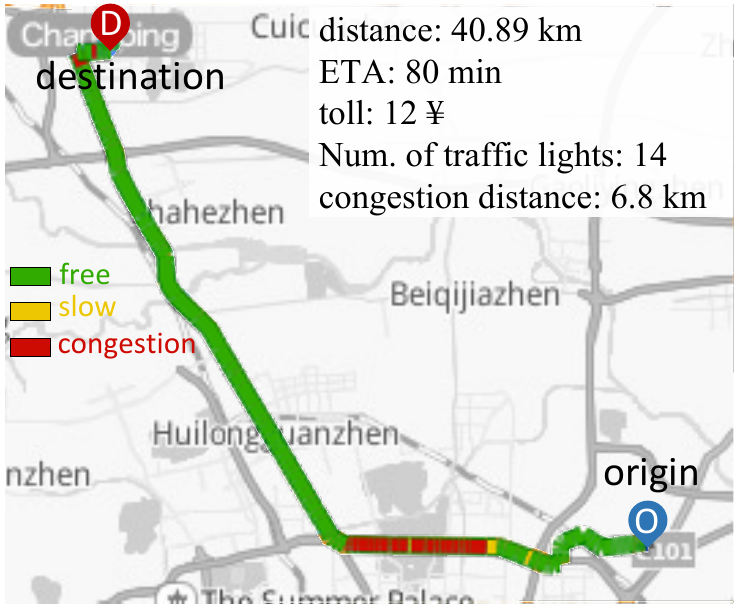}   %以pic.jpg的0.5倍大小输出
    \end{minipage}
    \label{figure:case_dist}
  }
  \subfigure[ST and the base model] %第二张子图
  {
    \begin{minipage}[t]{0.3\linewidth}
    \centering      %子图居中
    \includegraphics[width=\linewidth]{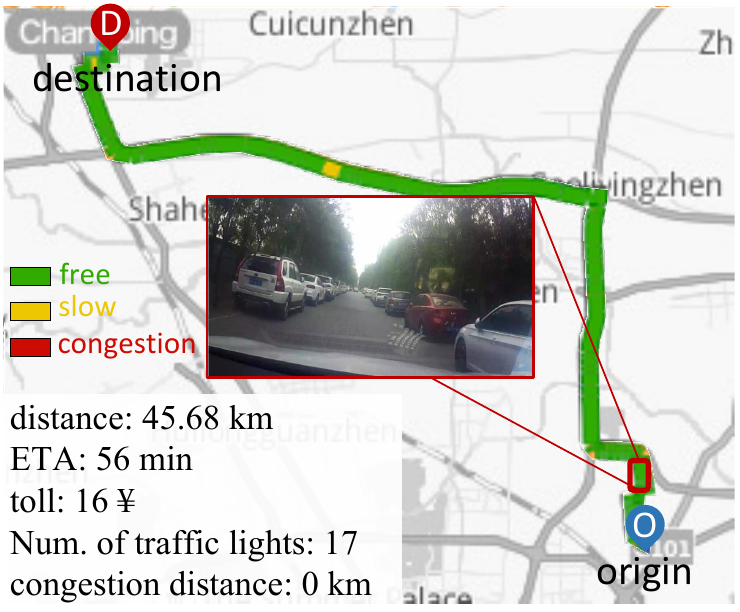}   %以pic.jpg的0.5倍大小输出
    \end{minipage}
    \label{figure:case_eta}
  }
  \subfigure[R4-C] %第三张子图
  {
    \begin{minipage}[t]{0.3\linewidth}
    \centering      %子图居中
    \includegraphics[width=\linewidth]{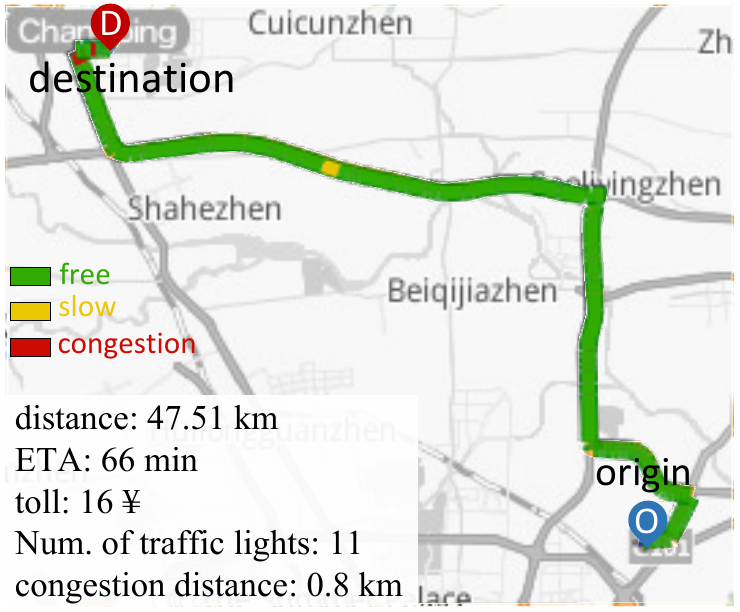}   %以pic.jpg的0.5倍大小输出
    \end{minipage}
    \label{figure:case_embedding}
  }

  \caption{\textbf{Cases of different route ranking models.} } %  %大图名称
  \label{figure:ranking_cases}  %图片引用标记
\end{figure*}

\subsection{Experiments on Model Clipping}

To speed up model inference, we employ ResNet20 instead of ResNet50 in online experiment, and the response time (RT) reduces from 100 ms to 60 ms. ResNet20 is able to capture information of 13 links, which means it can characterize route with length of about 2km since the average length of link is about 160m. It is sufficient for learning local features of route. Besides, as shown in Table \ref{table:comparison}, ResNet20 has similar performance with ResNet50.

\subsection{Online Experiments}

We conduct online experiments of several baseline models and R4, to manifest the performance of R4 in route recommendation:

\begin{itemize}
  \item \textbf{SD.} Recommend routes with the shortest distance.
  \item \textbf{ST.} Recommend routes with the shortest ETA. The ETA calculation leverages a regression model with real travel time as labels, which implies that ETA is a weighted combination of multiple indicators such as distance, realtime traffic condition, realtime road condition, etc.
  \item \textbf{ST-H.} Recommend routes with the shortest historical time-cost in statistics. Each link has an average statistical time-cost. The time-cost of route is a cumulation of link time-costs.
  \item \textbf{Base Model}. Predict DR of each route with the base model (Section \ref{section:basemodel}), and recommend routes with the lowest predicted DR.
  \item \textbf{R4-C}. Predict DR of each route with R4-C (Section \ref{section:methods}), and recommend routes with the lowest predicted DR.
\end{itemize}

Experiments are conducted in Beijing. Table \ref{table:online} provides experimental results of DR. Compared with the base model, R4-C decreases DR by 2.4$\%$, which indicates higher satisfaction of users. It manifests R4-C has better performance in DR prediction and route recommendation. Besides, in contrast with the base model and R4-C, SD, ST, and ST-H all have higher DR, implying lower satisfaction of users. 

The statistical features of routes recommended by different models are listed in Table \ref{table:online}. Comparing R4-C with the base model, the number of traffic lights and the number of turns have very small changes, which proves that routes recommended by R4-C provide sufficient intersections for users to turn. That is to say, the decrease of DR is not achieved by recommending routes with less intersections for users to turn. Moreover, the highway distance is reduced, which also eliminates the probability that R4-C decreases DR by inducing users to drive highways and consequently restricting users from turning into other routes. Route distance of SD is the shortest as expected, but it has much more traffic lights and much longer ETA than others. ST has shortest ETA but longest route distance and highest toll. ST-H has no obvious advantages.

Figure \ref{figure:ranking_cases} provides an example of recommended routes from same origin to same destination of different ranking models. Figure \ref{figure:case_dist} is the route recommended by SD. It has shortest distance but 7 km of congestion. In this case, ST-H recommend the same route with SD. Both ST and the base model recommend a route with barely no congestion as shown in Figure \ref{figure:case_eta}. In contrast with the route recommended by SD, it reduces time by 24 min, which seems to be a good choice. However, as shown in the picture, it has over 1 km of minor road with many cars parked on both sides of it, which is very hard to drive through and may give user a pretty bad experience. R4-C solves the problem as given in Figure \ref{figure:case_embedding}. It avoids the bad minor road above and chooses a major road instead at the beginning of the route.
It further proves that R4-C indeed captures implicit properties of routes and ranking routes by predicted DR of R4-C is able to avoid bad roads.

In online service, user embeddings are pre-calculated and updated daily. Online DR prediction retrieves them directly from database.  Since the framework supports parallel computing, it is possible to apply it to a large road network with hundreds of millions of links.

\section{Conclusion}
In this paper, a novel framework R4 is proposed to learn route representation and user representation in route recommendation task. For the accuracy of route representation, we denote route as a sequence of links, which includes both road segment information and turn information. The sparse $\&$ dense network we proposed obtains two types of route embeddings. One contains implicit characteristics of routes, the other contains local features of routes. For user representation, a sequence of user historical navigation is utilized to extract user preference. Moreover, a similarity unit is designed to reveal hidden features captured by link ID embeddings. It is helpful for discovering new attributes and correcting attributes errors, which in return benefits the learning of route representation. Ablation study proves the effectiveness of each unit in R4. Furthermore, the online experiment on Amap, the top-tier LBS-service provider in China, manifests the remarkable performance improvement of R4 for route recommendation.

\bibliographystyle{coling}
\bibliography{coling2020}

\end{document}